\documentclass[11pt]{llncs}
\usepackage[a4paper,twoside=false,top=2.32in,bottom=2.69in,left=2.25in,right=2in]{geometry}
\usepackage{times}
\usepackage{latexsym}
\usepackage{xcolor}
\usepackage{cite}
\usepackage{soul}
\usepackage{epigraph}
\usepackage{hyperref}
\usepackage{breqn}
\usepackage{versions}\excludeversion{ignore}\excludeversion{comment}
\title{Metaphor Interpretation\\ Using Word Embeddings\thanks{This paper was presented at the \textit{Conference on Intelligent Text Processing and Computational Linguistics (CICLing)} in March 2018 (Hanoi, Vietnam).}}
\author{Kfir Bar\inst{1} \and 
Nachum Dershowitz\inst{2} \and 
Lena Dankin\inst{2}\thanks{This work was carried out in partial fulfillment of the requirements for the Ph.D. degree of this author.}}
\institute{School of Computer Science, The College of Management,\\ Rishon LeZion, Israel
\\\email{kfirb@colman.ac.il}
\and
School of Computer Science, Tel Aviv University,\\ Ramat Aviv, Israel
\\\email{\{nachum,lenadank\}@tau.ac.il}}
\begin{document} 
\maketitle
\begin{abstract}
We suggest a model for metaphor interpretation using word embeddings trained over a relatively large corpus. Our system handles nominal metaphors, like \textit{time is money}. It generates a ranked list of potential interpretations of given metaphors. Candidate meanings are drawn from collocations of the topic (\textit{time}) and vehicle (\textit{money}) components, automatically extracted from a dependency-parsed corpus. 

We explore  adding  candidates derived from word association norms (common human responses to cues). Our ranking procedure considers  similarity between  candidate interpretations and  metaphor components, measured in a semantic vector space. Lastly, a clustering algorithm removes semantically related duplicates, thereby allowing other candidate interpretations to attain  higher rank. 

We evaluate using different sets of annotated metaphors, with encouraging preliminary results.
\end{abstract}

\epigraph{
Writing about metaphor is dancing with your conceptual clothes off, the innards of your language exposed by equipment more powerful than anything operated by the TSA. Still, one would be a rabbit not to do it in a world where metaphor is now top dog, at least among revived rhetorical devices with philosophical appeal.} 
{Carlin Romano,
``What’s a Metaphor For?'',\\
\textit{The Chronicle of Higher Education} (July 3, 2011)}

\section{Introduction}

Metaphor is pervasive in language and thought \cite{gentner_metaphor_2001}. 
Based on a quantitative analysis, Krennmayr \cite{e6cb9e2375544ca3a0ce48024803f20c} found that even in academic papers almost every fifth word is part of a metaphorical concept, broadly construed.

Already Aristotle  analyzed and wrote about the use of metaphor. ``Metaphor'', he says in the \textit{Poetics}, ``consists in giving the thing a name that belongs to something else.'' \textit{The sunset of life} is one of his examples.  
In \textit{Rhetoric} he explains:
``A simile is also a metaphor; for there is little difference: when the poet says, `He rushed as a lion,' it is a simile, but `The lion rushed' would be metaphor [`lion' referring to a human hero]; since both are brave.''

Following a definition provided by \cite{mar2008grammar}, 
A {\it metaphor} is a rhetorical figure, which is a peculiar expression of a sentiment different from the ordinary way. A {\it simile} is that figure by which we compare one object with another. 
In other words, a metaphor is a simile without any formal comparison \cite{mar2008grammar}.
Other examples of simile are: {\it as sweet as pie} (nominal) and {\it eat like a bird} (verbal); the expressions {\it life is a roller coaster}, {\it time is money}, and {\it you are my sunshine}, are nominal metaphors.

Aristotle heaps praise on metaphor:
\begin{quote}
To be a master of metaphor is the greatest thing by far. It is the one thing that cannot be learnt from others, and it is also a sign of genius.

Metaphor especially has clarity and sweetness and strangeness.

Words which make us learn something are most pleasant.\dots It is metaphor, therefore that above all produces this effect;
for when Homer calls old age stubble, he teaches and informs us through the genus;
for both have lost their bloom. 
\end{quote}

 Lakoff and Johnson \cite{COGS:COGS134} claim that the human conceptual system is extremely metaphorical in nature. They talk about that {\it metaphorical concepts} are being defined in terms of {\it nonmetaphorical concepts}. They explain that nonmetaphorical concepts are those that emerge directly from experience and can be defined in their own terms. Therefore, metaphorical concepts are composed of their own terms as well as terms of other concepts. By way of example, they mention the metaphor {\it time is money}; {\it money} is a limited resource, and limited resources are valuable. Therefore, {\it time} is valuable.
Generally speaking, they argue that most of the metaphorical concepts are abstract (e.g.\@ {\it time}, {\it emotions}, {\it ideas}), and that they are usually described metaphorically by concrete objects (e.g.\@ {\it food}, {\it physical objects}). Ortony et al.~\cite{ORTONY1978465} add that when using a metaphor the writer's goal is to convey only the metaphorical concept.

Metaphors are often used for expressing emotions, as a tool for visualizing  concepts.  A {\it broken heart} describes a sad feeling  caused by someone or something; it is not meant literally. It creates an image of a heart that is broken into pieces for conveying an extreme feeling of sadness. 
In \cite{mohammad2016metaphor}, it was shown that metaphors carry significantly  more emotions than do literal expressions. 
This is one of the reasons for metaphor being a useful device in creative expression. For example, it allows a writer to describe a concept that is difficult to explain directly through a creative emotional imagery. In \cite{lakoff1987}, {\it image metaphors} are defined as metaphors that map conventional mental images onto other conventional images with similar characteristics, as for example, describing a politician as a ``bulldozer''. This opens up many possibilities for creativity in writing.

A specific metaphor sometimes has an ambiguous interpretation. For example, when we say {\it memory is a river}, both {\it fluid} and {\it long} might be considered acceptable interpretations \cite{Roncero2015}. It has been shown in experiments that sentential context, too, may affect the meaning of the metaphor \cite{ORTONY1978465}.
The emotional characteristic of metaphor  increases the level of ambiguity, as people might interpret emotions in multiple ways. 

The rhetorician, I. A. Richards \cite{richards1965philosophy}, decomposes  a metaphor into two main components: the {\it tenor} and the {\it vehicle}. The tenor, or {\it topic}, is  that which is being described by potential meanings, referred to as \textit{properties}, of the vehicle. There are several metaphorical syntactic constructions. Similarly to other works on this topic, we focus on Noun-Noun constructions, that is,
metaphors of the form {\it Noun} is [a] {\it Noun}; {\it time is money}, for example. The first noun is the topic and the second, the vehicle. This type of metaphor is known as {\it nominal}.
Noun-Noun constructions may  extend  beyond two nouns. For example, Albert Einstein once said: ``All religions, arts, and sciences are branches of the same tree'', suggesting that the three topics are related.
\begin{ignore}In addition, to demonstrate the robustness of our approach, we experiment with Verb-Object constructions, for example {\it reflect concern}. \end{ignore}

The meaning of a metaphor may be related more to the topic, the vehicle, or to both in the same level. For example, when one says that {\it Joe is a chicken}, the meaning is usually  described as being {\it afraid}, which is more closely related to the vehicle {\it chicken} than to the topic {\it Joe}. On the other hand, Bob Dylan said in an interview on 1965, ``Chaos is a friend of mine'', a metaphor that can be interpreted as something {\it chaotic}, which is more related to the topic.

We describe a system that is designed for interpreting nominal metaphors, given without context. Similarly to previous works, we exploit a large corpus of text documents for semantically describing words and properties using a mathematical device. We use a word-embedding representation for calculating similarity between a candidate interpretation and the topic and the vehicle, so as to rank  candidates based on a semantic score. As a final step, we automatically cluster  results and keep only the best interpretations out of each cluster.

To summarize this paper's contributions:
\begin{enumerate}
  \item We provide a new and improved dataset.
  \item We extend previous works in this field using a richer semantic model for interpreting metaphors, and obtain competitive results.
  \item We show that clustering and filtering the results to leave only the best in each cluster improves performance.
  \item We show that using word associations as interpretation candidates, combined with collocations, improves performance, as do topic interpretations.
  \item We suggest some additional metrics for evaluation, such as mean reciprocal rank and mean average precision. And we use word senses (WordNet synonyms) for matching.
\end{enumerate}

The next section cites some related work. Our contributions and the results of experiments are described in the following two sections. Some conclusions are drawn in the final section.

\section{Related Work}
Different tasks relate to automatic metaphor processing. One is about automatically identifying metaphor in running text, that is, tagging words as being part of a metaphor or not. 
Many studies handle this. For example, Turney et al.~\cite{Turney:2011:LMS:2145432.2145511} automatically tag words in a given context as either {\it literal} or {\it metaphorical}, by training a supervised classifier. They focus on features that measure the level of abstractness of the word's context. They were able to show state-of-the-art performance on a dataset of adjective-noun metaphors (e.g.\@ {\it sweet child\/}).
For more information on metaphor identification, we refer the reader to \cite{shutova2016design}, a recent review of metaphor processing systems.

Before that, \cite{Birke06aclustering} presented a system for identifying literal and nonliteral usages of verbs, focusing on identifying metaphorical meaning, through statistical word-sense disambiguation and clustering algorithms. At a high level, they use a small set of manually sense-annotated sentences. Given a verb with its sentential context, they calculate the similarity between the input sentence and the annotated set, and decide on the sense that mostly occur within the most similar annotated sentences.

Neuman et al.~\cite{neuman2013metaphor} extended previous work~\cite{Turney:2011:LMS:2145432.2145511}, covering metaphors formed of only concrete concepts, by identifying {\it selectional preference} violations. A selection preference is a concept presented in \cite{light2002statistical}, claiming that words mentioned literally in a sentence, usually co-occur with word that belong to a {\it selected} semantic concept. They treated a violation of this idea as an indication for the nonliteral class.

The 
 computational task in which we are more interested, is \emph{interpretation}, interpreting a given metaphor. 
This very challenging task has garnered interest over the past few years. {\it Metaphor Magnet} \cite{Veale:2012:SVI:2390470.2390472}, allows users to enter a metaphor or simile, potentially augmented with sentiment polarity (e.g.\@ $+/-$); for example, {\it life is a $+$game}, including a {plus} sign for {\it game}, indicating a positive sentiment. Using sentiment this way allows users to provide some information about the context. To interpret a metaphor, the system expands the topic and vehicle with some corpus-based {\it stereotypes}, and then with the stereotype's properties. The properties that saliently occur with both, the topic and the vehicle, are returned as results. For Metaphor Magnet, a stereotype is a word that describes the topic/vehicle. The stereotypes and properties are discovered using Google n-grams, as it contains n-grams of the form ``X is a Y'' that help one understand how X is typically being described.

There are a few works that treat the text components as vectors of a higher dimension in a semantic space. This opens the possibility of using mathematical tools to calculate the similarity of two components, through measuring the distance between their corresponding vectors. Kintsch~\cite{kintsch2000metaphor} uses latent semantic analysis (LSA) \cite{Deerwester90indexingby} for modeling the vector space. They generate term vectors that highly correlate with both, the topic and the vehicle; correlation is measured by cosine similarity over the LSA vectors.  Metaphor interpretation is represented by the centroid vector of the most similar terms, and it does not necessarily represent a real word. 
Terai and Nakagawa \cite{terai2012corpus} use the same algorithm, over a slightly different semantic model. They use probabilistic latent semantic indexing (PLSI) \cite{hofmann1999probabilistic}, for finding potential properties, limiting to adjectives and verbs. We go down the same path, in the sense that we use a semantic model for calculating a score for the candidate properties. Similarly, we focus on adjectives and verbs as the only possible interpretations. Terai and Nakagawa also extended their process with a recurrent neural network trained over the properties and scores for finding the dynamic interaction between the properties.

The most relevant work for us is {\it Meta4meaning} \cite{meta4meaning},  an interpretation system for nominal metaphors. This work uses an LSA along with two  dimensionality reduction techniques, Singular Value Decomposition (SVD) and Non-negative Matrix Factorization (NMF). It only considers abstract words as candidate properties. The properties are ranked according to their association strength with both,  topic and  vehicle. It uses different aggregation methods for combining the association scores of the topic and the vehicle. The system shows a strong performance advantage over the human-annotated dataset provided by \cite{Roncero2015} compared with other systems.

Following {\it Meta4meaning}, we build a word-embedding model instead of LSA. Specifically, we use a 300-dimensional \textsf{GloVe} model \cite{pennington2014glove}. Word embeddings, specifically of the type that we are using, outperform SVD for analogy tasks~\cite{levy2014neural}. Since our task is more similar to analogy  than to  word similarity, we were led to believe that word embeddings may improve performance of  metaphor interpretation.

\section{Metaphor Processing}
Given a metaphor, we begin by generating a list of interpretation candidates. We do this by finding  collocations of the topic and vehicle individually, and consider each one of them as a potential candidate. For each candidate $c$, we calculate a topic semantic score, which is the cosine similarity between $c$ and the $k$ most significant collocations of the topic  ($k$ is a parameter) and aggregate it into a single score by averaging all scores. Similarly, we calculate a vehicle semantic score. 

In the next step, we calculate two pointwise mutual information (PMI) values, between $c$ and the topic and vehicle respectively. We add the frequency of $c$ as another score and combine all the five score functions in a log-linear structure, with weights assigned to each. The weights are adjusted automatically, as we describe in the following section. 

To remove semantically related interpretations from the list, we cluster the results and keep only the highest ranked candidates in each cluster. The remaining candidates are ranked according to their final score and the best $n$ candidates ($n$, too, is a parameter) are returned as interpretations. 

We now describe each step in greater detail.

\subsection{Potential Interpretations}
In our work, similar to other relevant works, e.g.\@ \cite{meta4meaning,Roncero2015}, a metaphor \textit{interpretation} is composed of a single word that conveys the main concept of the metaphor. For example, among the interpretations of the metaphor {\it city is a jungle} one can find {\it crazy} and {\it crowded}. It is natural to assume that an interpretation should be of a class of {\it describing} words, that is, words that are used for describing objects. Therefore, similar to other related works \cite{Veale:2012:SVI:2390470.2390472,meta4meaning}, we consider all adjectives as potential interpretations. In addition, we add verbs with an {\it ing} ending as candidates. In \cite{meta4meaning}, they only consider abstract words as potential interpretations. The level of abstractness of a word was measured by Turney et al.~\cite{Turney:2011:LMS:2145432.2145511} automatically for about 11,000 words. To avoid the limitation in using such a list, we did not go that route; we believe that most of the potential interpretations are adjectives.

\subsection{Dependency-Based Collocations}
Our interpretation process begins with extracting collocations of the vehicle and the topic individually using a relatively large corpus. Specifically, we use DepCC,\footnote{\url{https://www.inf.uni-hamburg.de/en/inst/ab/lt/resources/data/depcc.html}} a dependency-parsed ``web-scale corpus'' based on Common Crawl.\footnote{\url{http://commoncrawl.org}} There are 365 million documents in the corpus, comprising about 252B tokens. Among other preprocessing steps, every sentence was given with word dependencies discovered by MaltParser \cite{Nivre05maltparser:a}. We only use a fraction of the corpus containing some 1.7B tokens.
Here, we consider as collocation words that are found to be dependent in either the topic or the vehicle, and assigned with a relevant part-of-speech tag: adjective or verb+{\it ing}. The main assumption  is that many   potential modifiers of a given noun will appear somewhere in the corpus as a dependent in the dependency graph.

For example, the dependency-based collocations for  {\it school} are: {\it high}, {\it elementary}, {\it old}, {\it grad}, {\it middle}, {\it med}, {\it private}, {\it attending}, {\it graduating}, {\it secondary}, {\it leaving}, and {\it primary}. 

To eliminate noisy results that might transpire given that the corpus was generated from the open web, we preserve only candidates that have an entry in WordNet \cite{fellbaum98wordnet}.

\subsection{Word Association}
In parallel with our objective data-driven collocation extraction process, we experimented with word associations as an alternative, more subjective, process for generating interpretation candidates.  Word-association norms are repositories of pairs of words and their association frequency in a given population. The first word is a cue or trigger given to participants, and the second is the reported  associated word that first came to a subject's mind. For example, {\it bank} is paired with {\it money}, because the cue {\it bank} often elicits the response {\it money}. Those pairs form various semantic-relation types; some might not be deemed symmetric. Word association norms have been used in psychological and medical research, as well as a device for measuring creativity.

We use the University of South Florida (USF) free  association norms \cite{nelson2004university}\footnote{\url{http://w3.usf.edu/FreeAssociation}} for generating alternative candidates. This repository contains 5,019 cue words that were given to 6,000 participants beginning in 1973. We utilize this repository by adding all the associated words of the topic and vehicle individually. In this case, we allow words of all parts of speech to be considered as candidates. For example, the associations for cue  {\it school} are: {\it work}, {\it college}, {\it book}, {\it bus}, {\it learn}, {\it study}, {\it student}, {\it homework}, {\it teacher}, {\it class}, {\it education}, {\it USF} (!), {\it hard}, {\it boring}, {\it child}, {\it house}, {\it day}, {\it elementary}, {\it friend}, {\it grade}, {\it time}, {\it yard}.
We evaluate our system's performance with and without the associations; results are reported below. 

\subsection{Calculating Semantic Scores}
For each candidate we calculate a couple of semantic scores, one for the topic and one for the vehicle. We use word embeddings to transform every word into a continuous vector that captures the meaning of the word, as evidenced in the underlying corpus. 
We used pre-trained \textsf{GloVe} \cite{pennington2014glove} vectors; specifically, we use the ones that were trained over a 6B token corpus, comprising 400K vectors, each of 300 dimensions. 
In what follows, we denote the vector of a word $v$ by $w_v$.

We believe that the most significant collocations of the topic/vehicle tend to reliably represent the way the topic/vehicle, respectively, can be described in different contexts. Therefore, the semantic score ${\it sem}(c, t)$ of a candidate $c$ and the topic $t$ is the average cosine similarity between $w_c$ and the vectors of the $k$ most significant collocations of $t$. Similarly, ${\it sem}(c, v)$ is the semantic score of a candidate $c$ and the vehicle $v$. We experimented with different values for $k$. Results are reported in the next section.

\subsection{Final Scores}
For each candidate $c$, we calculate ${\it npmi}(c, t)$ and ${\it npmi}(c,v)$, the normalized pointwise mutual information (PMI) values for the topic and vehicle, respectively. Normalized PMI is similar to PMI, except that it is normalized between $-1$ and 1. The PMI between a candidate $c$ and a noun $n$ is calculated over the dependency graph; that is, we calculate the chances of seeing $c$ as a dependent of $n$ in a dependency graph.
We add ${\it freq}(c)$, the frequency of $c$, as another score, calculated over the entire corpus. 

To summarize, given a candidate $c$, the full list of scores is 
\[
\langle \textit{sem}(c, t), \textit{sem}(c, v), \textit{npmi}(c, t),  npmi(c, v), \textit{freq}(c)\rangle
\]
combined using a log-linear structure, with each score  amplified by a weight:
\[\textit{FinalScore}(c) = \sum_{k=1}^5 \lambda_k \log \textit{score}_k\]
We automatically adjust these weights over a development set of metaphors and interpretations to optimize for recall, as explained below. As a result, each candidate is ranked according to its final score.

\subsection{Clustering}
Lastly, we  cluster the list of candidates as a way to deduplicate it. We run clustering using word vectors for finding groups of words that have a strong semantic association of any kind, keeping only the best candidates in each cluster.

We use density-based spatial clustering of applications with noise (DBSCAN) for clustering. This  method  groups together vectors that are bundled in the space by forcing a minimum number of neighbors. Vectors that do not have the requisite number of neighbors, or in other words occur in low-density areas, are reported as noise and are not placed under any cluster. For us it means that they were not connected with other vectors, so they might have a unique meaning among the listed candidates. We treat such vectors as if they  form  singletons.

For example, among the interpretation candidates for the metaphor {\it anger is fire} we find {\it red} and {\it black}. After clustering, {\it black} is removed. As another example, the following candidates for the metaphor {\it a desert is an oven} may be grouped together: {\it eating}, {\it healthy}, {\it delicious}, {\it fried}, {\it spicy}, {\it leftover}, {\it veggie}, {\it steamed}, {\it lentil}, {\it roasted}, {\it homemade}, {\it yummy}, {\it creamy}, {\it glazed}, {\it seasoning}, {\it crunchy}, {\it baking}. (These likely result from the frequent misspelling of ``dessert'' in the corpus used.)

There are two parameters that need to be configured for DBSCAN: (1) $\varepsilon$  -- the radius of the consideration area around every vector; and 
(2) $\mu$  -- the minimum number of neighbors required in the consideration area. The distance measure should also be configured. We use the common Euclidean distance, which usually shows good performance in a relatively low-dimension space like ours.
Below we describe our experimental results, using different values for both parameters.
Table \ref{tab:res_ex} shows a few outputs for three different metaphors.

\begin{table}[t]
\caption{Results for several metaphors.}
\centering
\begin{tabular}{|l|l|l|}
\hline
\bf friendship &  \bf god is a & \bf typewriter is a \\
\bf is a rainbow & \bf fire & \bf dinosaur\\
\hline
\hline
{\it beautiful} & {\it burning} & {\it prehistoric}\\
{\it wonderful} & {\it fighting} & {\it fossilised}\\
{\it colorful} & {\it holy} & {\it extinct}\\
{\it forming} & {\it sacred} & {\it resembling}\\
{\it pink} & {\it good} & {\it feathered}\\
{\it great} & {\it absolute} & {\it robotic}\\
{\it bright} & {\it powerful} & {\it stuffed}\\
{\it magical} & {\it cannon} & {\it primitive}\\
{\it deep} & {\it dangerous} & {\it preserved}\\
{\it double} & {\it killing} & {\it gigantic}\\
{\it happy} & {\it almighty} & {\it antique}\\
{\it featuring} & {\it calling} & {\it lumbering}\\
{\it good} & {\it great} & {\it basal}\\
{\it vibrant} & {\it heavy} & {\it ancient}\\
{\it glorious} & {\it alive} & {\it oversized}\\
\hline 
\end{tabular}
\label{tab:res_ex}
\end{table}

\section{Experimental Results}

\subsection{Evaluation Set}
We evaluate our system with the dataset published by \cite{Roncero2015}, containing 84 unique topic/vehicle pairs that were associated with interpretations by twenty different study participants. Each participant was asked to assign interpretation for different aspects of the pairs, such as treating a pair as a metaphor (e.g.\@ {\it knowledge}/{\it power}, from the phrase {\it knowledge is power}) or as a simile (e.g.\@ {\it knowledge}/{\it power}, from {\it knowledge is like power}). We focus on the interpretation of metaphors, both lexicalized and non-lexicalized.

As a preprocessing step, we lemmatize the interpretations, so as to allow our method's results and the true interpretations to match more smoothly. Additionally, we allow interpretations to match if they are considered as synonyms in WordNet. In this work we focus on nominal metaphors, and since our collocation as well as word-embedding models were trained to handle unigrams, we had to modify some of the metaphors that have multiword vehicles; such multiwords are modified into a single words by eliminating the space characters, knowing it may cause performance reduction; For example, {\it sermon is a sleeping pill} is modified to {\it sermon is a sleepingpill}.

Each metaphor might be associated with more than one interpretation. As do other related works \cite{meta4meaning,Roncero2015}, we only consider interpretations that were assigned by at least five participants; we call them {\it qualified interpretations}. This leaves us with only 76 qualified metaphors (i.e.\@ metaphors with at least one qualified interpretation), with two qualified interpretations per metaphor on average. Table \ref{tab:evalset_ex} shows a few examples of interpretations  as assigned by 20 human annotators for the dataset of \cite{Roncero2015}.

\begin{table}[t]
\caption{Topic/vehicle pairs and associated properties.}\label{tab:evalset_ex}
\begin{center}
\begin{tabular}{|l||l|}
\hline
\bf Topic/Vehicle Pair &  \bf Associated Properties\\
\hline
\hline
{\it Skating}/{\it Flying} & {\it Free}; {\it Fast}; {\it Relaxing} \\
\hline 
{\it Store}/{\it Zoo} & {\it Crowded} \\
\hline 
{\it Wisdom}/{\it Ocean} & {\it Vast}; {\it Huge} \\
\hline 
{\it Job}/{\it Jail} & {\it Boring} \\
\hline 
\end{tabular}
\end{center}
\end{table}

\subsection{Evaluation Method}
To stay in line with related works \cite{meta4meaning}, we report  {\it Recall} @$K$, which is the average percentage of human-associated interpretations that are found in the top $K$ results. For example, the following results were generated for the pair {\it skating/flying} from Table \ref{tab:evalset_ex}: {\it incredible}, {\it high}, {\it free}, {\it great}, {\it fast}. Therefore, Recall@3 is 33\%, while Recall@5 is 66\%. We compare our results with \cite{meta4meaning}, which was evaluated on the same dataset following a similar preprocessing step. Therefore, we report on Recall at their reported $K$'s: 5, 10, 15, 25, and 50.

To measure the false positives reported by the system, we evaluate the results with two additional standard metrics: mean reciprocal rank (MRR) and mean average precision (MAP).

\subsection{Tuning System Weights}
Our log-linear structure is composed of a set of weighted score functions. We adjust the scores using a tuning process over a development set, composed of about 50\% of the metaphors. For each weight, we explore a range of possible scores, while we test all possible score combinations taking the brute force approach. For all scores except {\it freq}, we consider the range $0.1 \mathrel{..} 1$; because of scale differences, for {\it freq} we consider the range $1 \mathrel{..} 10$. 

As mentioned, we use DBSCAN to cluster the list of candidates so as to remove some of the semantically related ones. 
We take a similar brute force approach for tuning the DBSCAN parameters, $\varepsilon$ and $\mu$. We also tune $n$, the number of top results  taken from  each cluster. For tuning, we use the same development set, evaluated over MRR, MAP and Recall @$K$ values. Table \ref{tab:tuningParams} shows the ranges and best values of all the parameters we tune. 

We see that both semantic scores get higher weights than the npmi scores, suggesting that the semantic distance as measured by cosine similarity between the vectors of the candidates and the collocations of the topic/vehicle, is effective. The DBSCAN parameters are less stable across different metric optimizations. One thing we learn is that when optimizing for larger values of $@K$, DBSCAN requires dense areas around clustered vectors, resulting in a lower number of clusters. Additionally, the system does not benefit from high values of the DBSCAN $n$ parameter. It turns out that it is better to consider only one interpretation from each cluster.

\begin{table}[t]
\caption{System parameters tuned to maximize MRR, MAP and Recall@$K$. The second column shows the range of values considered.}\label{tab:tuningParams}
\begin{center}
\begin{tabular}{|l|c|c|c|c|c|c|c|c|}
\hline
\bf Parameter & \bf Range &  \multicolumn{1}{c|}{\bf MRR} & \multicolumn{1}{c|}{\bf MAP} & \multicolumn{1}{c|}{\bf @5} & \multicolumn{1}{c|}{\bf @10} & \multicolumn{1}{c|}{\bf @15} & \multicolumn{1}{c|}{\bf @25} & \multicolumn{1}{c|}{\bf @50}\\\hline\hline
DBSCAN $\varepsilon$ & $1   \mathrel{..} 6$ & 4 & 5 & 4 & 5 & 5 & 4 & 4\\
\hline 
DBSCAN $\mu$ & $1  \mathrel{..} 5$ & 4 & 1 & 6 & 1 & 5 & 5 & 5\\
\hline 
DBSCAN $n$ & $1  \mathrel{..}  12$ & 1 & 1 & 1 & 1 & 1 & 1 & 1\\
\hline 
$\textit{sem}(c, t)$ & $0.1  \mathrel{..}  1$ & 0.6 & 0.6 & 0.6 & 0.1 & 0.1 & 0.6 & 0.6\\
\hline 
$\textit{sem}(c, v)$ & $0.1  \mathrel{..}  1$ & 1.1 & 1.1 & 1.1 & 0.6 & 0.6 & 1.1 & 1.1\\
\hline 
$\textit{npmi}(c, t)$ & $0.1  \mathrel{..}  1$ & 0.1 & 0.1 & 0.1 & 0.1 & 0.1 & 0.1 & 0.1\\
\hline 
$\textit{npmi}(c, v)$ & $0.1  \mathrel{..}  1$ & 0.1 & 0.1 & 0.1 & 0.1 & 0.1 & 0.1 & 0.1\\
\hline 
$\textit{freq}(c)$ & $1  \mathrel{..}  10$ & 3 & 3 & 5 & 7 & 7 & 5 & 3\\
\hline 
\end{tabular}
\end{center}
\end{table}

\subsection{Evaluation Results}
We evaluate our system against the 76 ``qualified'' metaphors in the dataset. For each metaphor, our system generates the top 100  interpretation results, which are then compared with the metaphor's human-associated qualified interpretations. For the clustering parameters and scoring weights, we use the tuned values  reported in the previous subsections. Since we tune  for different evaluation metrics, here we individually use each set of values for generating the top 100 results and calculating MRR, MAP and recall at all the relevant $K$ values.
Table \ref{tab:results} summarizes the evaluation results at MRR, MAP and Recall@5, @10, @15, @25, and @50, for each set of parameter values. We observe that when optimizing the system for Recall@50 we  at least get close to the best result for all other evaluation metrics. Therefore, in what follows we use  the parameter values optimized for @50.

\begin{table}[t]
\caption{Each row shows evaluation results when using optimal parameter values for the metric mentioned in the first column.}\label{tab:results}
\begin{center}
\begin{tabular}{|l|r|r|r|r|r|r|r|r|}
\hline
 \bf Optimization &  \multicolumn{1}{c|}{\bf MRR} & 
 \multicolumn{1}{c|}{\bf MAP} & \multicolumn{1}{c|}{\bf @5} & \multicolumn{1}{c|}{\bf @10} & \multicolumn{1}{c|}{\bf @15} & \multicolumn{1}{c|}{\bf @25} & \multicolumn{1}{c|}{\bf @50}\\\hline\hline
MRR & \bf{0.312} & \bf{0.170} & 0.198 & 0.254 & 0.278 & 0.405 & 0.562\\
\hline 
MAP & \bf{0.312} & \bf{0.170} & 0.198 & 0.254 & 0.278 & 0.405 & 0.562\\
\hline 
@5 & 0.302 & 0.166 & \bf{0.207} & 0.258 & 0.270 & 0.430 & 0.548\\
\hline 
@10 & 0.233 & 0.151 & 0.180 & \bf{0.273} & 0.322 & 0.374 & 0.521\\
\hline 
@15 & 0.245 & 0.160 & 0.151 & 0.262 & \bf{0.331} & 0.392 & 0.513\\
\hline 
@25 & 0.302 & 0.166 & \bf{0.207} & 0.258 & 0.270 & \bf{0.430} & 0.548\\
\hline
@50 & \bf{0.312} & \bf{0.170} & 0.198 & 0.254 & 0.278 & 0.405 & \bf{0.562}\\
\hline 
\end{tabular}
\end{center}
\end{table}

We compare our results with the ones reported by Meta4meaning \cite{meta4meaning}, evaluating over the same set of metaphors and following similar preprocessing steps. Table \ref{tab:compResults} compares the results reported by both systems. While our system somewhat underperforms for the lower values of Recall $@k$, it is doing slightly better on @25 and @50. These results show that, while our system has a better overall coverage,  correct interpretations are concentrated more in the lower part of the ranked list that we produce. With more work, we expect to be able to filter out many of the non-associated interpretations, thereby ranking the correct ones higher in the list.

\begin{table}[t]
\caption{Comparison with Meta4meaning.}\label{tab:compResults}
\begin{center}
\begin{tabular}{|l|r|r|r|r|r|r|r|r|}
\hline
 \bf System &  \multicolumn{1}{c|}{\bf MRR} & \multicolumn{1}{c|}{\bf MAP} & \multicolumn{1}{c|}{\bf @5} & \multicolumn{1}{c|}{\bf @10} & \multicolumn{1}{c|}{\bf @15} & \multicolumn{1}{c|}{\bf @25} & \multicolumn{1}{c|}{\bf @50}\\\hline\hline
Meta4meaning & N/A & N/A & \bf 0.221 & \bf 0.303 & \bf 0.339 & 0.397 & 0.454 \\
\hline 
Ours & 0.312 & 0.170 & 0.198 & 0.254 & 0.278 & \bf 0.405 & \bf 0.562 \\
\hline 
\end{tabular}
\end{center}
\end{table}

To measure the effect of clustering on the results, we evaluate our system running with and without clustering. When running with clustering, we use the optimized set of parameters, as reported in Table \ref{tab:tuningParams}. Table \ref{tab:clusteringCompareResults} compares our system's results, with and without clustering. We learn that when using clustering, our system was able to eliminate noise in lower parts of the ranked list of candidates, thereby making room for alternative and correct interpretations that ranked lower without clustering.

\begin{table}[t]
\caption{Evaluation results, with and without clustering.}\label{tab:clusteringCompareResults}
\begin{center}
\begin{tabular}{|l|r|r|r|r|r|r|}
\hline
 \bf Method & \multicolumn{1}{c|}{\bf @5} & \multicolumn{1}{c|}{\bf @10} & \multicolumn{1}{c|}{\bf @15} & \multicolumn{1}{c|}{\bf @25} & \multicolumn{1}{c|}{\bf @50}\\\hline\hline
w/o clustering & 0.198 & 0.254 & 0.278 & 0.351 & 0.534 \\
\hline 
w/ clustering & 0.198 & 0.254 & 0.278 & \bf 0.405 & \bf 0.562 \\
\hline 
\end{tabular}
\end{center}
\end{table}

\begin{figure}[t]
\begin{center}
\includegraphics[scale=0.70]{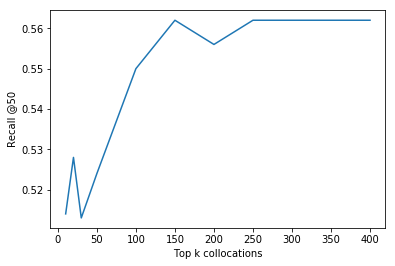}
\end{center}
\caption{Evaluation results as Recall@50, measured over different $k$ values for the maximum number of collocations we take from the topic/vehicle for calculating the semantic score.}
\label{fig:kValues}
\end{figure}

Recall that our topic/vehicle semantic scores are defined as the cosine similarity between the candidate vector and the top $k$ collocations of the topic/vehicle. We tested our system with different $k$ values; Figure \ref{fig:kValues} shows evaluation results as Recall@50 when running the system with different $k$ values. Observe that it gets maximized at higher values of $k$, suggesting that the meaning of the topic/vehicle is usually more complex, and that it takes multiple properties to describe when comparing it vis-\`a-vis  candidate interpretations.

Finally, we check how our system's performance is affected by adding word associations as an additional source for generating interpretation candidates. 
When we run our system using only dependency-based collocations as candidates, we obtain Recall@50 score of 0.551. This was improved to 0.562 when we add word associations as candidates.

\subsection{Improving the Dataset}
Overall, the system could not generate even one correct interpretation (among its 50 best results) for 20 out of the 76 evaluated metaphors. Some of those metaphors did not have a correct interpretation anywhere in the list, even beyond the best 50; for example, {\it music is a medicine}.
Taking a closer look at the dataset, we found that some metaphors did not come with any correct interpretation in its interpretation list, even when taking into account all the provided interpretations, not just qualified ones. For example, take the metaphor {\it education is a stairway}. The suggested interpretations are {\it higher}, {\it steps}, {\it upward}, {\it long}, {\it passage}, {\it ascension}, {\it climbing} -- none of which qualified. Most of these interpretations do not reflect the true meaning of this metaphor ({\it steps}, {\it passage} and {\it climbing} are themselves metaphors; {\it long} is surely not intended; {\it higher} and {\it upward} make little sense); we would rather suggest {\it enabling} as a more suitable interpretation. For {\it job is a jail}, the only qualified interpretation is {\it boring}, while the more accurate interpretation, {\it confining}, was proposed by fewer then 5 annotators, and therefore did not pass the bar. These are only a few of the examples that encouraged us to perform our own annotation process over the entire dataset. This was done by a native English speaker. We override the original interpretations with the newer ones, resulting in a slightly larger dataset, because with the new annotations some unqualified metaphors now qualify. 

In addition to these new annotations, we extended the dataset with 14 new metaphors  extracted from \cite{Lakoff91}, among them {\it words are weapons} and {\it logic is gravity}. We followed the same annotation process to assign interpretations for the new metaphors. The extended (and improved) dataset contains 98 metaphors with refined interpretations. The full list of modifications can be found in the dataset (published at \emph{to be supplied in the final version}).
We intend to extend it even further in the future.

Table \ref{tab:verbNounResults} compares evaluation results for the original and improved datasets. 
The degraded results we get for the latter is explained by the fact that, for most metaphors in the dataset, our improvement process removed the majority of suggested interpretations. Fewer human-annotated interpretations means fewer successful matches, making our improved dataset harder to interpret to begin with.

\begin{ignore}
\subsection{Evaluating on Verb-Object Metaphors}
To test the robustness of our technique, we also evaluate our system on a different type of metaphor: Verb-Object constructions, where the verb is used metaphorically. For example, the verb \textit{stir} is usually used to indicate some sort of motion, but in {\it stir excitement}, {\it stir} should be interpreted as {\it provoke} or {\it stimulate}. As opposed to the Noun-Noun construction, for the Verb-Object the interpretations are verbs, rather than adjectives. 

We experimented on a dataset provided by \cite{Shutova2010}, which consists of 41 Verb-Object metaphors. Out of those, five metaphors consist of phrasal verbs (e.g.\@ \textit{brush aside accusation}), and therefore were removed for the purpose of our experiments, as our system cannot currently deal with phrasal verbs. In addition to the Verb-Object metaphors, this dataset presents 11 Verb-Subject constructions, and some of the instances also consist of phrasal verbs. For example, \textit{report leaked} is such a construction, where  \textit{leaked} should be interpreted as \textit{disclosed}. Due to the small number of examples, we focus on Verb-Object examples only, and plan to revisit the Verb-Subject construction when we obtain substantially more instances of that type. This type of construction will require some minor modifications in the technique proposed for the interpretation of Verb-Object constructions.
\end{ignore}

\begin{table}[t]
\caption{Evaluation results when running on different datasets.}\label{tab:verbNounResults}
\begin{center}
\begin{tabular}{|l|r|r|r|r|r|r|r|r|}
\hline
 \bf Dataset &  \multicolumn{1}{c|}{\bf MRR} & \multicolumn{1}{c|}{\bf MAP} & \multicolumn{1}{c|}{\bf @5} & \multicolumn{1}{c|}{\bf @10} & \multicolumn{1}{c|}{\bf @15} & \multicolumn{1}{c|}{\bf @25} & \multicolumn{1}{c|}{\bf @50}\\\hline\hline
Original & 0.312 & 0.170 & 0.198 & 0.254 & 0.278 & 0.405 & 0.562 \\
\hline 
Improved & 0.151 & 0.073 & 0.051 & 0.070 & 0.114 & 0.171 & 0.311 \\
\hline 
\end{tabular}
\end{center}
\end{table}

\begin{ignore}
To support this new metaphor type, we had to slightly modify the collocation extraction process, so instead of considering adjectives and verb+{\it ing} that appear as dependent of the either the topic or the vehicle, here we consider all verb types that appear as a head of the noun object. 
\end{ignore}

\section{Conclusions}

      We have described a system that interprets nominal metaphors, provided without a context. Given a metaphor, we generate a set of interpretation candidates and rank them according to how strongly they are associated with the topic, as well as with the vehicle.  Candidates are generated using two techniques. First, we find collocations of the topic and  vehicle, focusing on adjectives as well as gerunds, which were found to be dependent of the topic/vehicle in at least one sentence in a large corpus. We add to that list word associations of both. This addition has proven effective.

Our ranking procedure combines a number of scores assigned for each candidate, which are based on normalized PMI as well as cosine similarity between the representing \textsf{GloVe} vectors of the candidates and the topic/vehicle collocations. The scores are aggregated using a weighted log-linear structure. We tune the weights automatically, optimizing for various evaluation metrics: MRR, MAP and Recall@$K$ for different $K$ values. We found that with small $K$, the similarity between  candidate and  topic becomes more important than  other score functions.
overl
In a post-processing step, we cluster the results using DBSCAN and keep only the best candidates out of each cluster. Our system benefits thereby.

Our system was evaluated against a set of metaphors that were assigned with properties by 20 human evaluators. We compare our results with Meta4meaning and obtained competitive results.

Additional work is needed to handle the cases mentioned in the analysis section, especially, cleaning the results from candidates that have an opposite meaning from the ones we are looking for.

Potential future directions include working on additional types of metaphors, as well as additional languages. We plan to improve the current evaluation technique; one option, which we're considering, is to measure the effect of metaphor interpretation on common NLP tasks, such as machine translation.
We will also be looking at the analysis of metaphors in context.

\subsection*{Acknowledgement}
This work was supported in part by the Blavatnik Family Foundation.

\bibliographystyle{splncs03}
\bibliography{metaphor}

\end{document}